\documentclass[letterpaper, 10 pt, conference]{ieeeconf}  \IEEEoverridecommandlockouts                              
\usepackage[utf8]{inputenc}
\usepackage[T1]{fontenc}

\usepackage{cite}
\usepackage{amsmath,amssymb,amsfonts}
\usepackage[ruled,vlined]{algorithm2e}
\SetKwInput{KwInput}{Input}                
\SetKwInput{KwOutput}{Output}              
\DeclareMathOperator*{\argmax}{arg\,max}
\DeclareMathOperator*{\mydeg}{deg}
\DeclareMathOperator*{\rand}{random.sample}

\usepackage{caption}
\usepackage[labelformat=simple]{subcaption}

\usepackage{float}

\usepackage{multirow}
\usepackage{graphicx}
\usepackage{textcomp}
\usepackage{hyperref}
\usepackage{xcolor}
\usepackage{tabularx,booktabs}
\newcolumntype{Y}{>{\centering\arraybackslash}X}

\usepackage{orcidlink}
\usepackage{balance}

\def\BibTeX{{\rm B\kern-.05em{\sc i\kern-.025em b}\kern-.08em
    T\kern-.1667em\lower.7ex\hbox{E}\kern-.125emX}}

\title{\LARGE \bf
GMC-Pos: Graph-Based Multi-Robot \\ Coverage Positioning Method
}

\author{Khattiya Pongsirijinda$^{1}$, Zhiqiang Cao$^{1}$, Muhammad Shalihan$^{1}$, Benny Kai Kiat Ng$^{1}$,\\ Billy Pik Lik Lau$^{1}$, \IEEEmembership{Member,~IEEE}, Chau Yuen$^{2}$, \IEEEmembership{Fellow,~IEEE}, and U-Xuan Tan$^{1}$, \IEEEmembership{Member,~IEEE}%
\thanks{$^{1}$K. Pongsirijinda, Z. Cao, M. Shalihan, B. K. K. Ng, B. P. L. Lau, and U-X. Tan are with the Engineering Product Development, Singapore University of Technology and Design, Singapore 487372 {(email: \{khattiya\_pongsirijinda, zhiqiang\_cao, muhammad\_shalihan\}@mymail.sutd.edu.sg, \{benny\_ng, billy\_lau, uxuan\_tan\}@sutd.edu.sg}).}%
\thanks{$^{2}$C. Yuen is with the School of Electrical and Electronic Engineering, Nanyang Technological University, Singapore 639798 (email: chau.yuen@ntu.edu.sg).}%
}

\begin{document}

\maketitle
\thispagestyle{empty}
\pagestyle{empty}

\begin{abstract}
Nowadays, several real-world tasks require adequate environment coverage for maintaining communication between multiple robots, for example, target search tasks, environmental monitoring, and post-disaster rescues. In this study, we look into a situation where there are a human operator and multiple robots, and we assume that each human or robot covers a certain range of areas. We want them to maximize their area of coverage collectively. Therefore, in this paper, we propose the \underline{G}raph-Based \underline{M}ulti-Robot \underline{C}overage \underline{Pos}itioning Method (GMC-Pos) to find strategic positions for robots that maximize the area coverage. Our novel approach consists of two main modules: graph generation and node selection. Firstly, graph generation represents the environment using a weighted connected graph. Then, we present a novel generalized graph-based distance and utilize it together with the graph degrees to be the conditions for node selection in a recursive manner. Our method is deployed in three environments with different settings. The results show that it outperforms the benchmark method by 15.13\% to 24.88\% regarding the area coverage percentage.
\end{abstract}

\section{Introduction}
\label{sect:I}

Area coverage has been playing a significant role in achieving various robotics-related tasks, such as path planning and area exploration. This topic is also critically important for search and rescue (SAR) operations \cite{rescue-survey2020}, such as post-disaster monitoring, awareness of victims' conditions, and target searching. At the same time, utilizing the robot as relaying nodes for wireless ad hoc networks has also become mainstream in recent years. According to current research, there is room to enhance how robots should efficiently locate in order to maximize the area coverage as much as possible.

\begin{figure}
  \begin{center}
  \includegraphics[width=\linewidth]{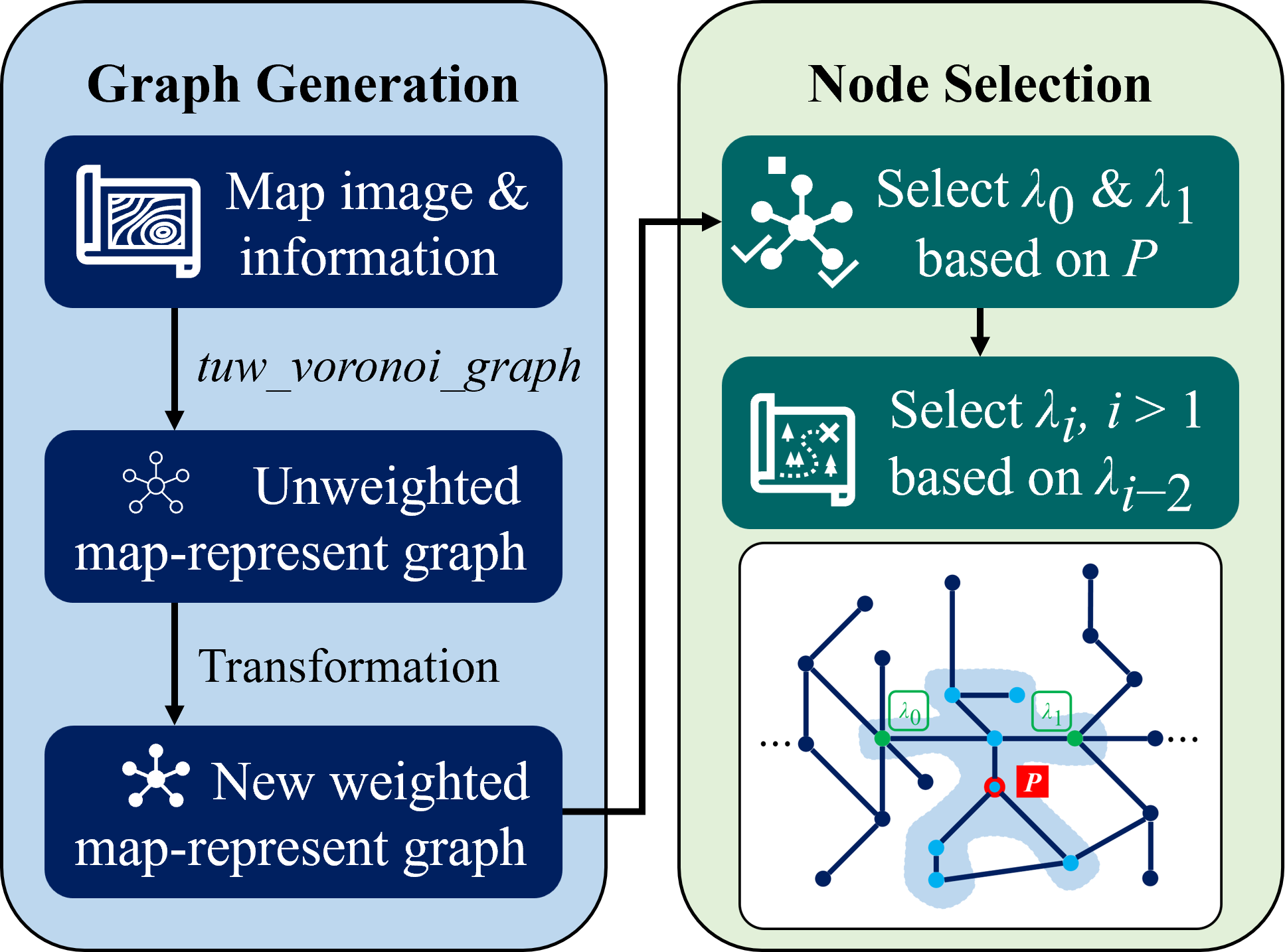}
  \vspace*{-3mm}
  \setlength{\belowcaptionskip}{-20pt}
  \caption{Overview of the GMC-Pos framework}
  \label{fig:overview}
  \end{center}
\end{figure}

According to the survey \cite{Survey2017}, currently, making the robots meet at their initial or some rendezvous positions is currently one of the most used assignments for rearranging the robot positions. Some strategies were proposed for different research objectives. There was a graph-based rendezvous \cite{rend-graph2012} proposed to be used together with exploration. Bio-inspired techniques, such as the ant algorithm \cite{rend-post2013} and bacterial chemotaxis \cite{rend2020}, were also applied for post-exploration meetings. Some works focused on connectivity-preserving \cite{rend-conn2017, rend-explo2022}, and communication-limited rendezvous \cite{rend-mission2022}. In spite of the fact that gathering robots by rendezvous techniques is systematic, it does not satisfy our primary goal, which is to cover the environment.

Although previous studies have not directly treated it in much detail, map-grid-based and graph-based strategies are the most popular approaches due to the benefits associated with map representation. The area coverage problem is often studied together with other topics. There are applications in various domains, such as path and motion planning \cite{voronoi2013, topo-map-pathplan2019, graph-pathplan2019, graph-motionplan2022, ac-graph-replan2023}, pathfinding \cite{graph-trian-pathfind2021}, multi-robot exploration \cite{graph-explo2021, graph-explo2022, ac-graph-patrol2010}, SAR tasks \cite{rescue2022}, or even strategic positioning for robot soccer teams \cite{pos-robocup2015, pos-robocup2015-2}. One of the related research matches our purposes and conditions. It is about the multi-robot coverage of a known environment \cite{ac-known-map2017}. Since, in our case, we have a map image as a prior before the positioning stage, it can also be considered a known environment. However, the existing methods that apply the map-grid-based approach \cite{pos-robocup2015-2, ac-known-map2017, ac-graph-replan2023} can have problems from unbalanced grid cell size and high computational time, especially when deploying a high number of robots. On the other hand, the existing graph-based methods utilize different types of graphs, for example, Voronoi diagrams \cite{pos-robocup2015, graph-pathplan2019, graph-explo2021, graph-motionplan2022}, Delaunay triangulation \cite{graph-trian-pathfind2021}, and bipartite graphs \cite{pos-robocup2015}, which are not aimed to be used for maximum area coverage purposes.

There also have been studies and applications of the Maximal Covering Location Problem (MCLP) \cite{MCLP_orig, MCLP1, MCLP2, MCLP3}, which is known to be NP-hard \cite{NPHard}. The mentioned problem aims to locate a number of facilities to maximize the amount of covered demand. For our problem studied in this paper, if we solve it in the same sense, we may consider the environment area as the demand and the robots as the facilities. However, our problem setting will have even more constraints than the standard MCLP \cite{MCLP_orig}. Firstly, the demand nodes were originally conceived to be sparse discrete points, but in our case, all map grids representing the area must be taken into account since our goal is to maximize the coverage area. Secondly, the mobile robots are different from the classical facilities in MCLP. Since the desired robot positions are also selected from the same map grids, they will simultaneously be demand nodes and facilities. Therefore, our unconventional problem requires a novel approach to solve it differently.

This paper proposes the \textbf{G}raph-Based \textbf{M}ulti-Robot \textbf{C}overage \textbf{Pos}itioning Method (GMC-Pos) to generate balanced and high coverage for 2D maps, which consists of two modules: graph generation and node selection.  Firstly, we present a graph generation method to construct the weighted connected graph to represent the environment. This graph behaves like the topological map of the environment, i.e., it will be according to the connectivity and structure of the environment. The generated graph nodes are placed in reachable areas and not obstructed by obstacles and walls. This is further enhanced by the property of the connected graphs, in which there is always at least one path between graph nodes. In addition, each edge of this map-represent graph is also assigned with its length as they will be beneficial for calculating the inter-position distances afterward. Secondly, we introduce a novel node selection strategy. In particular, the process is based on a recursive fashion with some graph-related requirements. Note that there are multiple robots and also one human operator in our setting, which can be treated like a robot as well, i.e., able to move around the environment. The robots will spread through the map according to the human position. In this module, we also construct a novel graph-based generalized distance combining the Euclidean distance and Dijkstra's shortest path length. This distance is used for all the relevant situations since it can realistically measure the distance between positions in the environment for both those that are and are not the graph nodes. Subsequently, as we aim to select the nodes as the robot positions for maximizing the area coverage, the nodes are chosen bidirectionally starting from the human position in the range of each robot area coverage radius. We then focus on the nodes with the highest degree in the range since they mostly represent the intersection or the centers of sub-areas. Finally, among these nodes, we select the one that is the furthest from the previous nodes in order to spread through the environment as much as possible. The main contributions of this paper are shown as follows:

\begin{itemize}
    \item To efficiently represent the environment for distance-calculated purposes, we propose a graph generation method to create a map-represent connected graph with the edge length as the weight.
    \item To obtain the maximum area coverage possible in any given environment, we propose a new recursive node selection strategy, which is based on the novel graph-based generalized distance and the graph nodes' degrees.
    \item We implement the GMC-Pos and test it in six scenarios using three different maps. Our method is compared with a benchmark to show its excellent performance in area coverage percentage.
\end{itemize}

The remainder of this paper proceeds as follows: Firstly, in Section \ref{sect:II}, the GMC-Pos is described in two subsections, namely graph generation and node selection. Next, in Section \ref{sect:III}, the details about the simulation settings, evaluation metric, and benchmark method will be provided. Then, the results will be presented and discussed in Section \ref{sect:IV}. Finally, in Section \ref{sect:V}, the main findings are concluded, and the directions of future research are addressed.

\section{GMC-Pos Positioning Method}
\label{sect:II}

\begin{figure}
  \begin{center}
  \includegraphics[width=\linewidth]{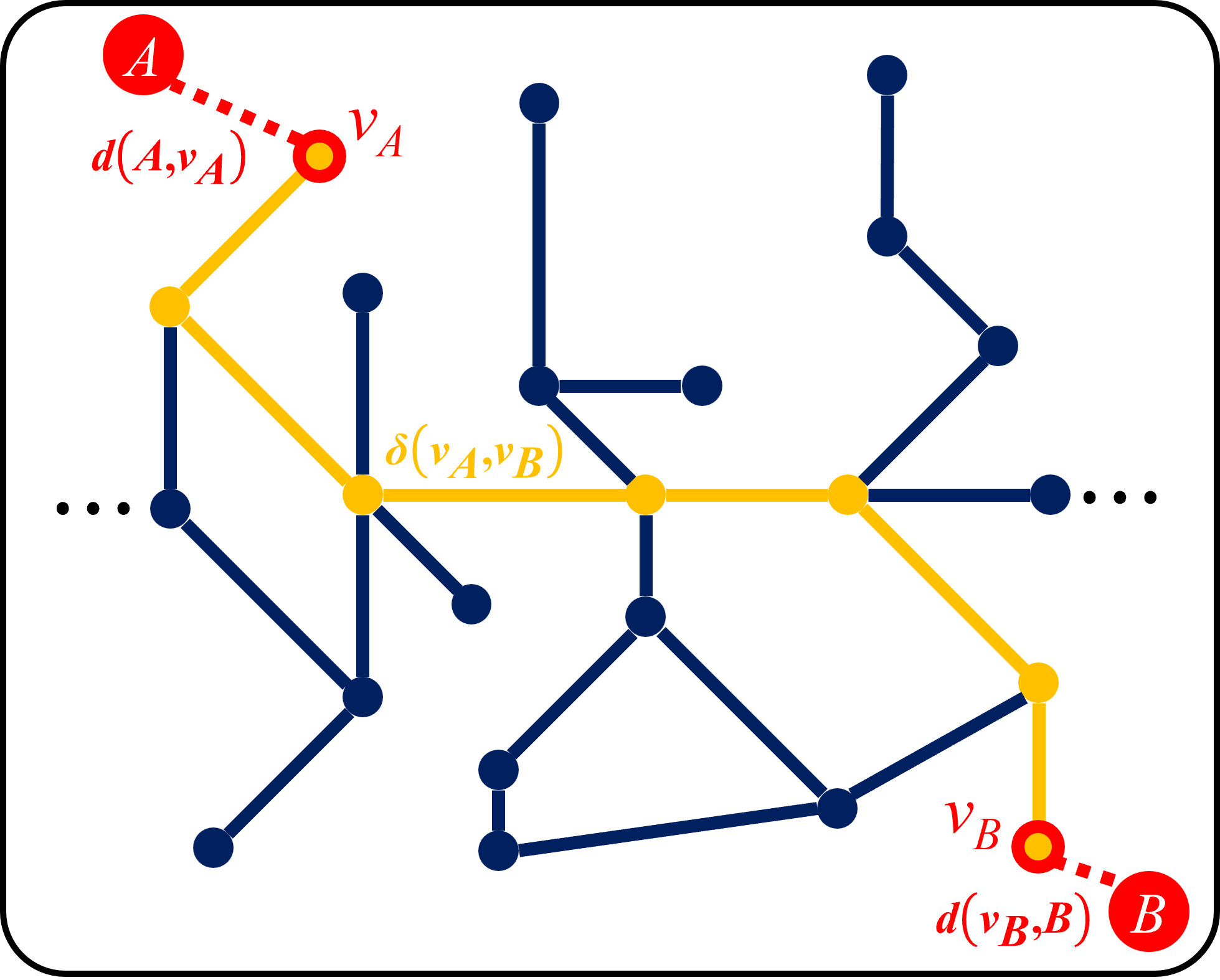}
  \vspace*{-3mm}
  \setlength{\belowcaptionskip}{-15pt}
  \caption{An example of the generalized graph-based distance from $A$ to $B$: $d_G(A,B)$}
  \label{fig:d_G}
  \end{center}
\end{figure}

This section describes the details of our proposed method, GMC-Pos, which consists of two modules: graph generation and node selection. Given a fully explored map, we have obtained map information. Then we can start generating a connected graph and select the appropriate nodes based on our novel strategy to be robot positions further. Each process will be explained in the following subsections.

\subsection{Graph Generation}

\begin{algorithm}
\DontPrintSemicolon
  \KwInput{The set of nodes: $V$, the number of robots: $N$, the scaled map height and width: $H$, $W$, the set of robot positions from a previous recursive step: $\Lambda$}
  \KwOutput{The set of robot positions $\Lambda$}
  {
    // Calculate $V_\eta$ \\
    $\alpha = \max(\{ H,W \})/N$ \\
    $V_\eta = \{ \}$ \\
    \For{$\lambda$ in $\Lambda\cup \{P\}$}{
        \For{$v$ in $V$}{
            \If{$d_G(v,\eta) < 2r$ and $d_G(v,\lambda) \geq \alpha$}{
                $V_\eta = V_\eta \cup \{v\}$            
            }
        }
    }
    // Calculate $V_\eta^*$ \\
    $V_\eta^* = \{ \}$ \\
    $maxdeg = \max_{v \in V_\eta}(\mydeg(v))$ \\
    \For{$v$ in $V_\eta$}{
        \If{$\mydeg(v)$ is equal to $maxdeg$}{
            $V_\eta^* = V_\eta^* \cup \{v\}$
        }
    }
    // Select the node as $\lambda_i$ and add it to $\Lambda$ \\
    $maxdist = \max_{v \in V_\eta^*}(d_G(v, \eta))$ \\
    \For{$v$ in $V_\eta^*$}{
        \If{$d_G(v)$ is equal to $maxdist$}{
            $\Lambda = \Lambda \cup \{v\}$ \\
            $V = V \setminus \{v\}$ \\
            \textbf{break}
        }
    }
 }
\KwRet{The set of robot positions $\Lambda$}
\caption{GMC-Pos's node selection process}
\label{algo:selection}
\end{algorithm}

The representing graph for the fully explored map is generated in the form of a connected graph by the Voronoi distillation \cite{voronoi2013}, which was implemented as a part of the ROS package, $tuw\_voronoi\_graph$ \cite{tuw2017, tuw2019}. We use the graph-based approach because considering all the positions on the map requires high computational time. Moreover, the main benefit of this graph is that it is connected and spanned thoroughly on the map. That means there is always at least one path between each graph node. The nodes also locate only in the explored area and do not overlap with obstacles. Thus, the graph nodes and edges are already sufficient for representing the environment. The graph generation can be customized by changing values of segment length, crossing optimization, and end segment optimization. However, since the unweighted graph created by the $tuw\_voronoi\_graph$ always uses the bottom left corner of the map image as the position $(0.0,0.0)$, we adjust the graph to have proper node coordinates for any map origin. Otherwise, the coordinate of each node will not be its actual position on the map. Moreover, as the distance between nodes will be presented in Section \ref{sect:node_selection}, we transform the graph into a weighted graph in the structure of a Python package, $NetworkX$ \cite{networkx}.

For the sake of convenience, each node in the generated graph is represented by its coordinate, while the distance between the connected nodes is assigned to be the weight of the corresponding edges. Therefore, let $G_0 = (V_0, E_0)$ be an unweighted graph generated by the $tuw\_voronoi\_graph$ where
\begin{equation}
    V_0 = \{ (x_v^0, y_v^0) \in \mathbb{R}^2 \}\text{.}
\end{equation}
We reconstruct $G_0$ into a weighted graph $G = (V,E)$. Let $(x_\text{map}, y_\text{map})$ be the map origin from the map image. The set of nodes can be denoted as follows:
\begin{equation}
    V = \{ (x_v^0 + x_\text{map}, y_v^0 + y_\text{map}) \in \mathbb{R}^2\} \text{ for all } (x_v^0, y_v^0) \in V_0\text{.}
\end{equation}
The edges in $E$ are still based on $E_0$, but are updated with the adjusted nodes in $V$. The weight of each edge $e=(u,v)\in E$ is assigned as follows:
\begin{equation}
    w(e) = d(u,v)\text{,}
\end{equation}
where $d$ is the Euclidean distance. This graph $G$ will be used afterward for node selection, which will be presented in Section \ref{sect:node_selection}.

\subsection{Node Selection}
\label{sect:node_selection}

Before looking into the selection process, we introduce a novel distance called the generalized graph-based distance. The generalized graph-based distance can be illustrated in Fig. \ref{fig:d_G}. {This distance is a better measurement than the Euclidean distance because it depends on the map-represent graph paths. So, it acts in accordance with the connectivity and structure of the map. We define it as follows:
\begin{equation}
    d_G(A,B) = d(A,v_A) + \delta(v_A,v_B) + d(v_B,B)\text{,}
\end{equation}
where $d$ is the Euclidean distance, $\delta$ is the length of Dijkstra’s shortest path \cite{dijkstra1959} calculated by a $NetworkX$ function, and $v_A$, $v_B$ are the nodes that are closest to $A$, $B$ in the Euclidean manner, respectively.

The generalized graph-based distance can realistically measure the distance between any point on the map for both the graph nodes and those that are not. For example, in the case of the distance between nodes $u$ and $w$, we have $u = v_u$ and $w = v_w$. Hence,
\begin{equation}
    d_G(u,w) = \delta(u,w)\text{,}
\end{equation}
which is just the length of Dijkstra’s shortest path between $u$ and $w$.

Moving to look at the node selection, the overall process is shown in Algorithm \ref{algo:selection}. We will choose the nodes as the robot positions based on the novel strategy in a balanced bidirectional manner to maximize the total area coverage. Let $\Lambda = \{ \lambda_0, \lambda_1, ..., \lambda_{N-1} \} \subseteq V$ be the set of nodes that are selected as the positions for $N$ robots. The human operator position $P$ and the robot area coverage radius $r$ are required as the input. The selection process is constructed using the recursion as follows:
\begin{align}
    \lambda_0 &= \argmax_{v\in V_P^*} \{ d_G(v,P) \}\text{,} \quad \lambda_1 = \argmax_{v\in V_P^*} \{ d_G(v,P) \}\text{,} \notag \\
    \lambda_i &= \argmax_{v\in V_{\lambda_{i-2}}^*} \{ d_G(v,\lambda_{i-2})\} \quad \text{if } i = 2,3, ..., N-1 \text{,} \label{eq:selection}
\end{align}
where
\begin{align}
    V_\eta^* &= \argmax_{v\in V_\eta} \{ \mydeg(v) \}\text{,} \label{eq:Vu*} \\
    V_\eta &= \{ v\in V | d_G(v,\eta) < 2r \land d_G(v,\lambda) \geq \alpha, \notag\\
    &\hspace{4ex} \forall\lambda\in\Lambda\cup \{P\}\}\setminus\Lambda\}\text{.} \label{eq:Vu}
\end{align}
Since the proposed selection strategy contains various novel components, it is important to clarify which each equation is used for which purposes. Starting from eq. (\ref{eq:Vu}), $V_\eta$ contains the nodes within $2r$. At the same time, the balance of nodes in $V_\eta$ spread thoroughly from all previously selected nodes in $\Lambda$ needs to be considered. Thus, we choose $\alpha$ as
\begin{equation}
    \alpha = \frac{\max(\{ H,W \})}{N}\text{,} \label{eq:alpha}
\end{equation}
where $H = H_0 \cdot Res$, $W = W_0 \cdot Res$, and $H_0$, $W_0$, $Res$ are the height, width, and resolution of the map, respectively.

Next, for eq. (\ref{eq:Vu*}), $V_\eta^*$ is a subset of $V_\eta$ in eq. (\ref{eq:Vu}), but it will contain only the nodes with the highest degree. We prefer these nodes because we can infer they have many connections in the map. Therefore, in most cases, they are the intersections or the centers of separate rooms, which will consequently affect the area coverage. Finally, the selection process (\ref{eq:selection}) is the last step after we have obtained the nodes with a maximum degree from eq. (\ref{eq:Vu*}). Among those nodes, we select the one that is the furthest from $\eta$ according to the bidirectional manner.

\section{Simulation}
\label{sect:III}
The details of the simulation and results of the GMC-Pos will be presented in this section. In the first subsection, we will describe how the simulations are conducted. Then, in the second subsection, we will introduce the evaluation metric. And finally, the method that we use as the benchmark will be explained in the third subsection.

\subsection{Simulation Setup}

\begin{figure}
     \centering
     \begin{subfigure}[b]{0.475\linewidth}
         \centering
         \includegraphics[width=\linewidth]{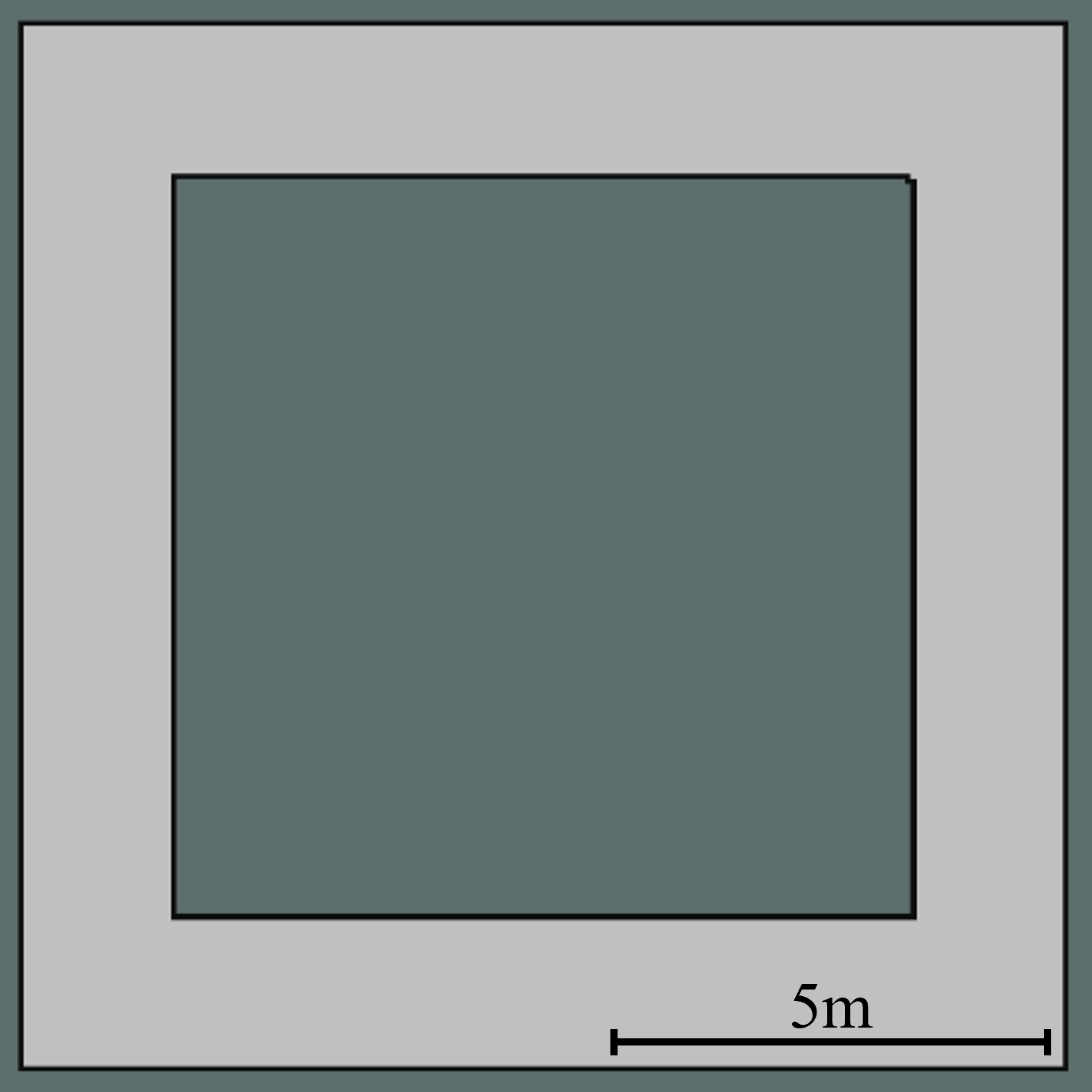}
         \caption{Map 1}
         \label{fig:map1}
     \end{subfigure}
     \begin{subfigure}[b]{0.4\linewidth}
         \centering
         \includegraphics[width=\linewidth]{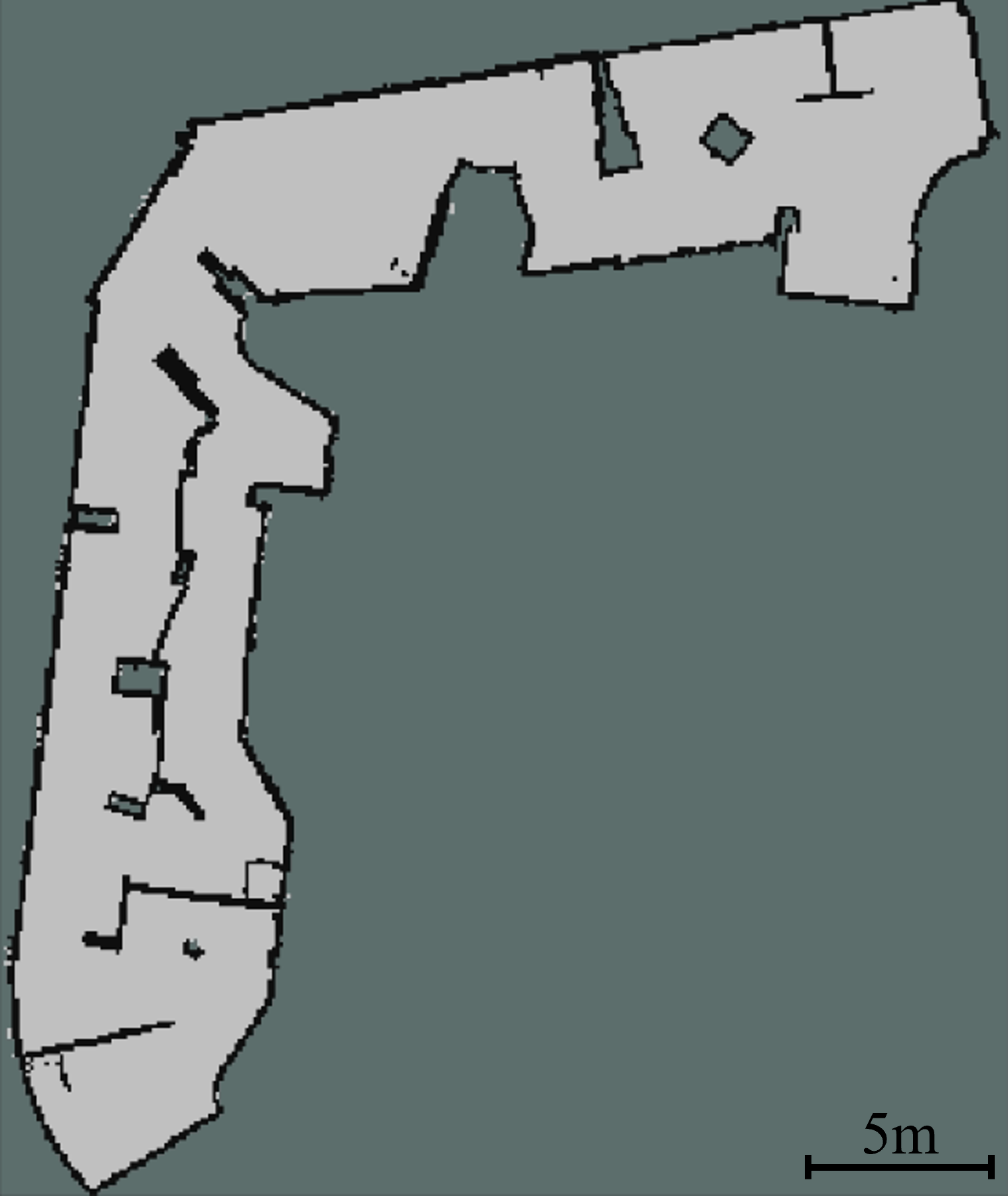}
         \caption{Map 2}
         \label{fig:map2}
     \end{subfigure}
     \begin{subfigure}[b]{0.9\linewidth}
         \centering
         \includegraphics[width=\linewidth]{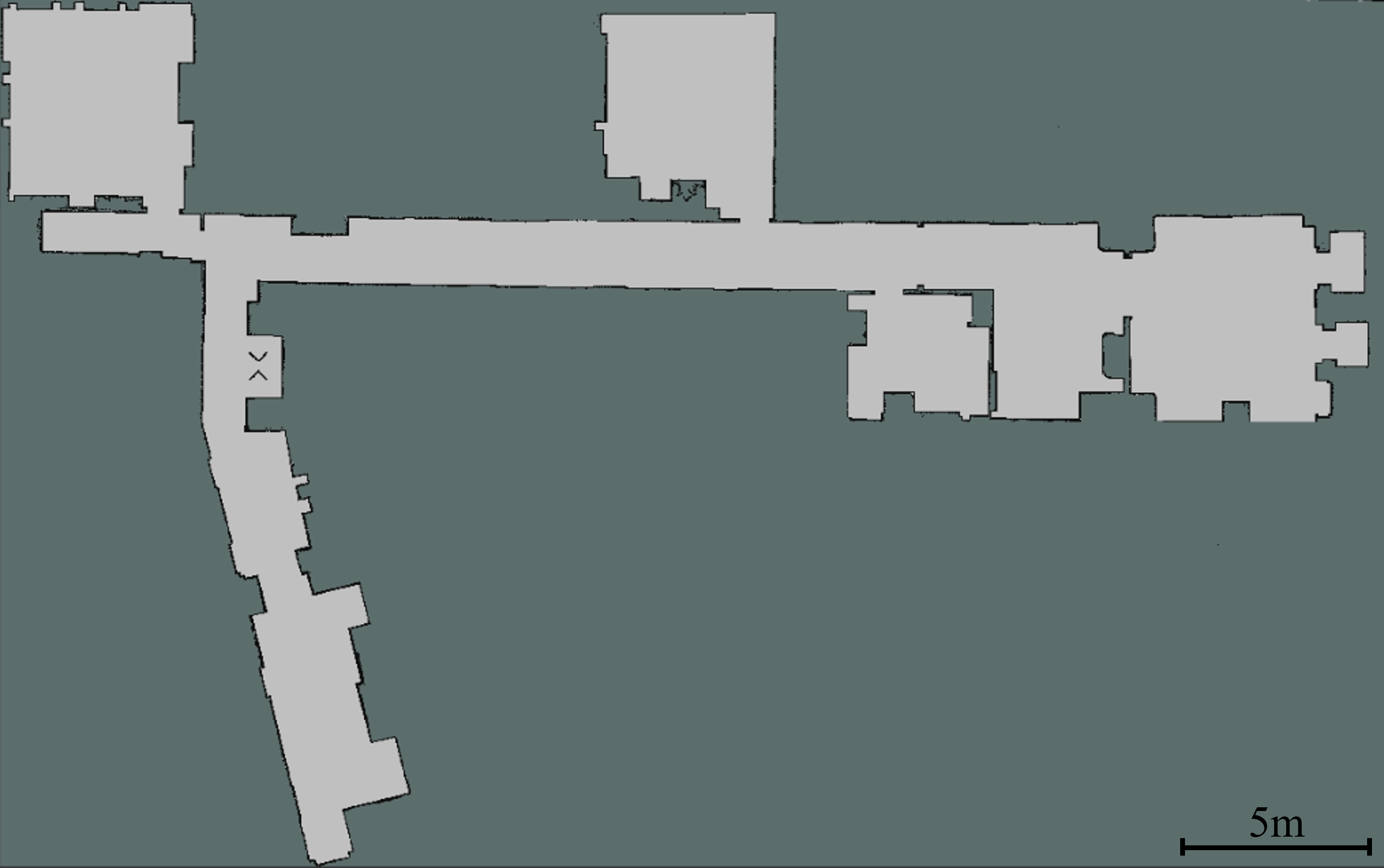}
         \caption{Map 3}
         \label{fig:map3}
     \end{subfigure}
        \caption{Environment maps}
        \label{fig:maps}
\end{figure}

\begin{table}[h!]
\centering
\begin{tabular}{|c|c|c|c|}
\hline
Scenario     & Map               & Number of Robots & Operator Location
\\ \hline
1A           & \multirow{2}{*}{Map 1} & 3             & Bottom left       \\ \cline{1-1} \cline{3-4} 
1B           &                        & 3             & Top right 
\\ \hline
2A           & \multirow{2}{*}{Map 2} & 5             & Center            \\ \cline{1-1} \cline{3-4} 
2B           &                        & 6             & Right side        \\ \hline
3A           & \multirow{2}{*}{Map 3} & 5             & Right side        \\ \cline{1-1} \cline{3-4} 
3B           &                        & 6             & Left side         \\ \hline
\end{tabular}
\caption{Simulation scenarios}
\label{tab:scenarios}
\end{table}

\begin{algorithm}
\DontPrintSemicolon
  \KwInput{The set of unoccupied occupancy grid cells: $O$, the number of robots: $N$, the scaled map height and width: $H$, $W$, the set of robot positions from a previous recursive step: $\Psi$}
  \KwOutput{The set of robot positions $\Psi$}
  {
    // Calculate $O_\phi$ \\
    $\alpha = \max(\{ H,W \})/N$ \\
    $O_\phi = \{ \}$ \\
    \For{$o$ in $O$}{
        \If{$\alpha \leq d(o,\phi) < 2r$}{
            $O_\phi = O_\phi \cup \{o\}$            
        }
    }
    // Select the node as $\psi_i$ and add it to $\Psi$ \\
    $\psi = \rand(O_\phi,1)$ \\
    $\Psi = \Psi \cup \{\psi\}$ \\
    $O = O \setminus \{\psi\}$ \\
 }
\KwRet{The set of robot positions $\Psi$}
\caption{Conditional Random's node selection process}
\label{algo:condrand}
\end{algorithm}

All simulations are conducted using ROS Melodic with Ubuntu 18.04 on a Desktop PC with Xeon(R) CPU E5-1680 v3 @ 3.20GHz×16 and 31.3 GB RAM. RViz is used for visualization. We perform multi-robot simulations with one human operator in three environments, as shown in Fig. \ref{fig:maps}. Map 1 is a loop corridor of size 12.20m$\times$12.20m, Map 2 is an actual indoor area of size 27.10m$\times$32.20m set up to be more complicated by using boxes and partitions, and Map 3 from \cite{tuw2017} is a floor map of size 37.37m$\times$23.38m containing long narrow corridors and rooms. Subsequently, we have the scenarios as shown in Tab. \ref{tab:scenarios}. The purpose of considering Scenarios 1A and 1B is to preliminarily determine if the GMC-Pos can accurately handle when the best robot positions are heuristically known. Meanwhile, the rest of the scenarios are mainly for testing the efficiency of GMC-Pos in various conditions, which have different numbers of robots and operator locations.

\begin{figure*}
     \centering
     \begin{subfigure}[b]{0.24\linewidth}
         \centering
         \includegraphics[width=\linewidth]{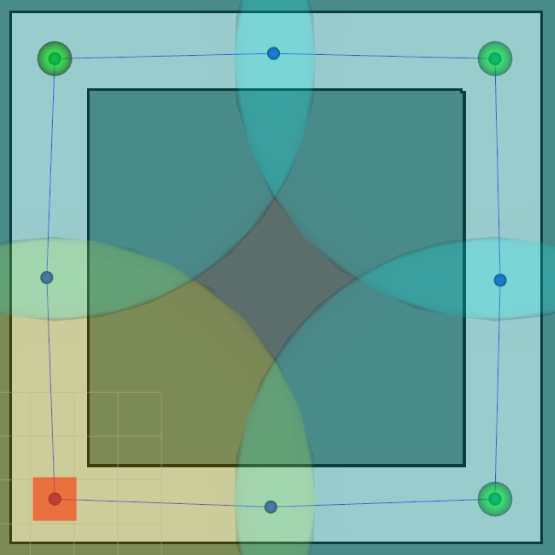}
         \caption{1A: GMC-Pos}
         \label{fig:1A-GMC-Pos}
     \end{subfigure}
     \begin{subfigure}[b]{0.24\linewidth}
         \centering
         \includegraphics[width=\linewidth]{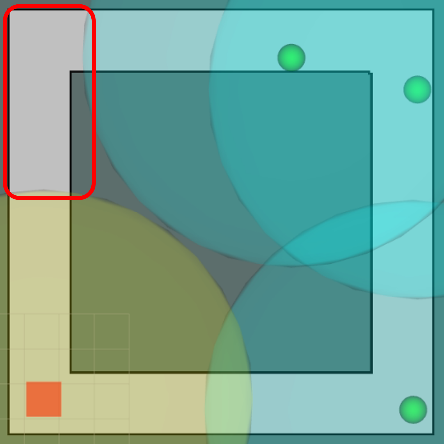}
         \caption{1A: Conditional Random}
         \label{fig:1A-CR}
     \end{subfigure}
     \begin{subfigure}[b]{0.24\linewidth}
         \centering
         \includegraphics[width=\linewidth]{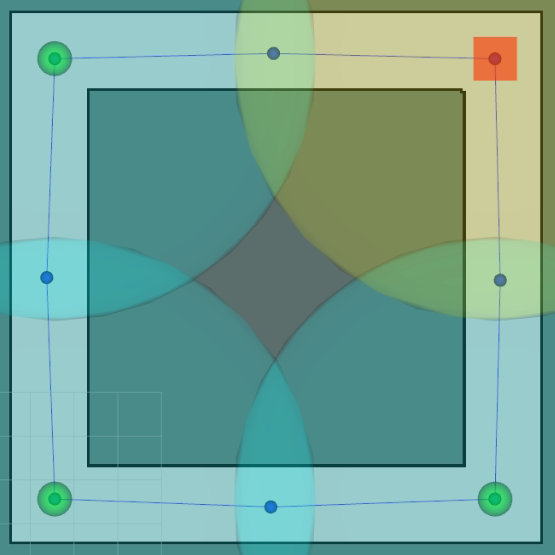}
         \caption{1B: GMC-Pos}
         \label{fig:1B-GMC-Pos}
     \end{subfigure}
     \begin{subfigure}[b]{0.24\linewidth}
         \centering
         \includegraphics[width=\linewidth]{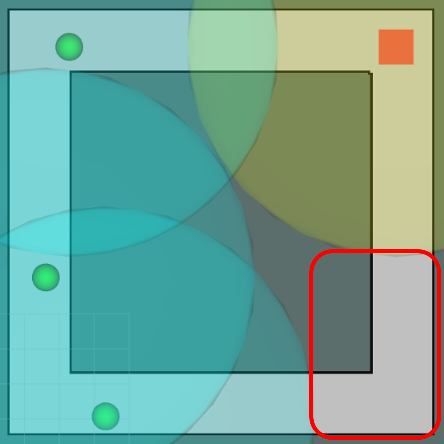}
         \caption{1B: Conditional Random}
         \label{fig:1B-CR}
     \end{subfigure} \\
     \begin{subfigure}[b]{0.24\linewidth}
         \centering
         \includegraphics[width=\linewidth]{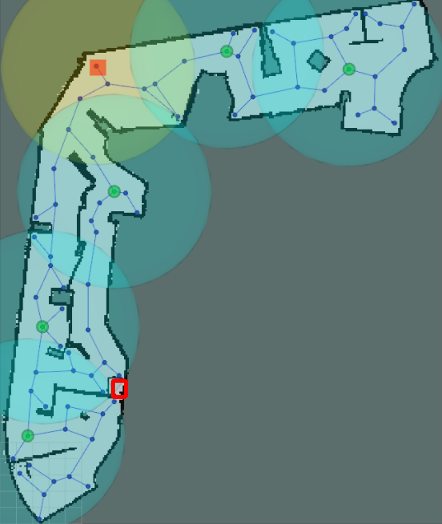}
         \caption{2A: GMC-Pos}
         \label{fig:2A-GMC-Pos}
     \end{subfigure}
     \begin{subfigure}[b]{0.24\linewidth}
         \centering
         \includegraphics[width=\linewidth]{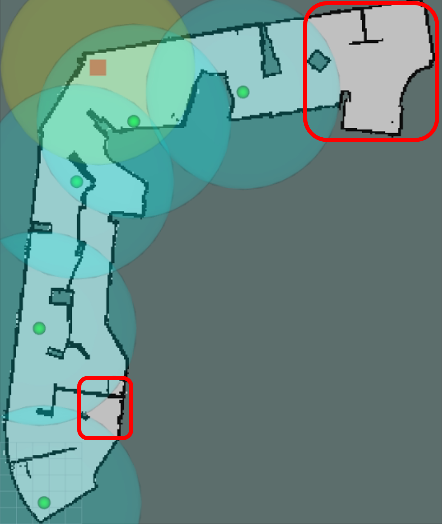}
         \caption{2A: Conditional Random}
         \label{fig:2A-CR}
     \end{subfigure}
     \begin{subfigure}[b]{0.24\linewidth}
         \centering
         \includegraphics[width=\linewidth]{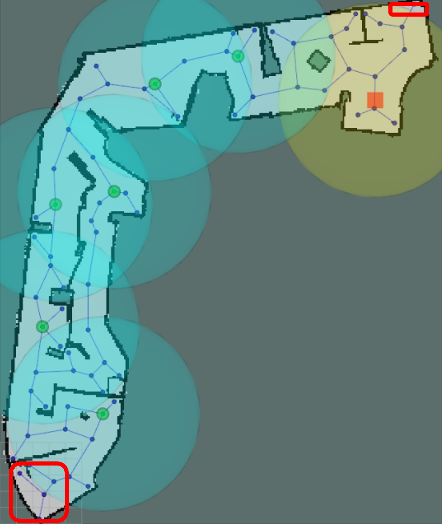}
         \caption{2B: GMC-Pos}
         \label{fig:2B-GMC-Pos}
     \end{subfigure}
     \begin{subfigure}[b]{0.24\linewidth}
         \centering
         \includegraphics[width=\linewidth]{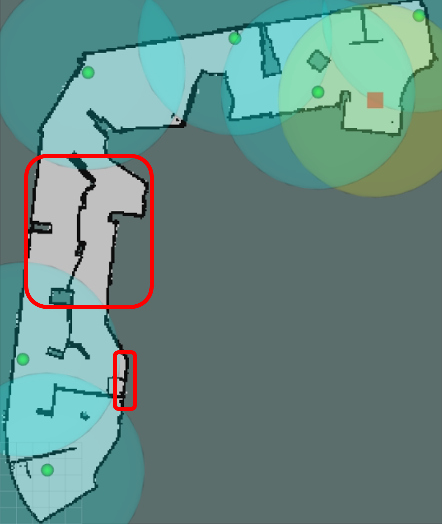}
         \caption{2B: Conditional Random}
         \label{fig:2B-CR}
     \end{subfigure} \\
     \begin{subfigure}[b]{0.24\linewidth}
         \centering
         \includegraphics[width=\linewidth]{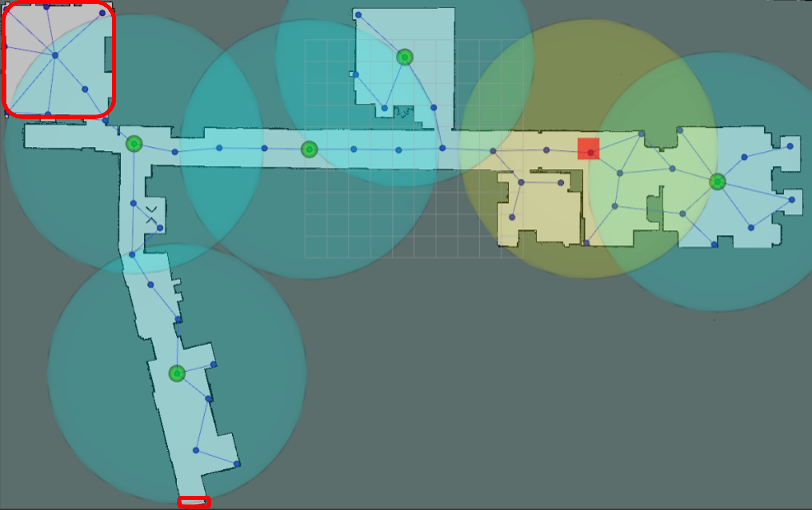}
         \caption{3A: GMC-Pos}
         \label{fig:3A-GMC-Pos}
     \end{subfigure}
     \begin{subfigure}[b]{0.24\linewidth}
         \centering
         \includegraphics[width=\linewidth]{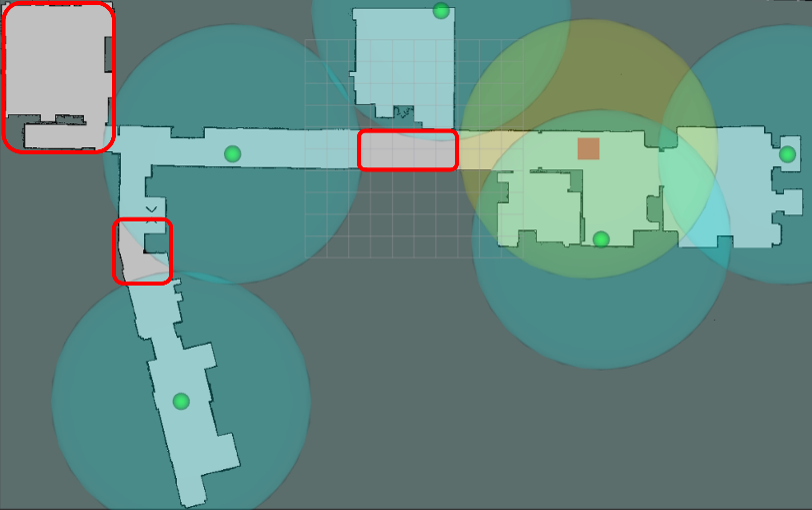}
         \caption{3A: Conditional Random}
         \label{fig:3A-CR}
     \end{subfigure}
     \begin{subfigure}[b]{0.24\linewidth}
         \centering
         \includegraphics[width=\linewidth]{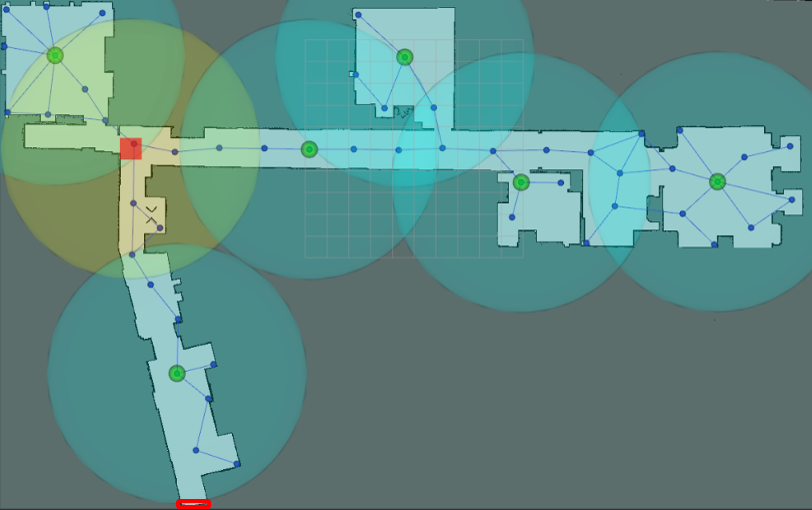}
         \caption{3B: GMC-Pos}
         \label{fig:3B-GMC-Pos}
     \end{subfigure}
     \begin{subfigure}[b]{0.24\linewidth}
         \centering
         \includegraphics[width=\linewidth]{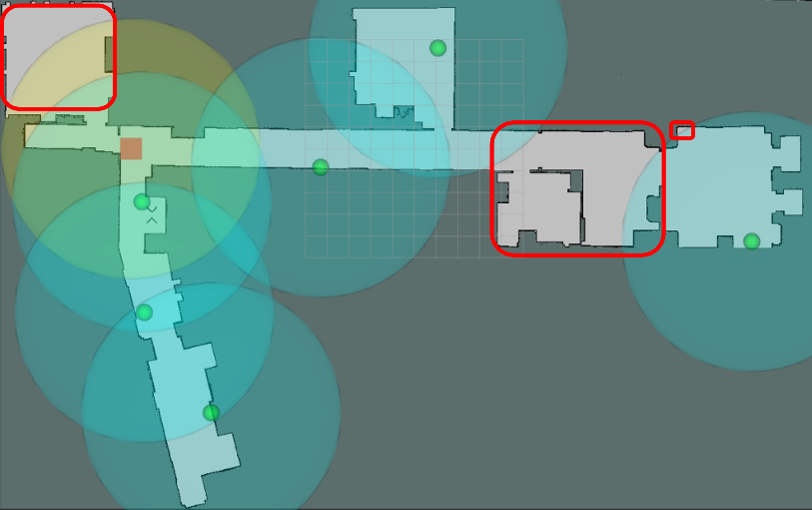}
         \caption{3B: Conditional Random}
         \label{fig:3B-CR}
     \end{subfigure}
        \setlength{\belowcaptionskip}{-5pt}
        \caption{Results of the GMC-Pos (the proposed method) and the Conditional Random. The green circles represent the robots, the red square represents the human operator, the blue translucent circles represent the areas covered by the corresponding robot, the yellow translucent circle represents the area covered by the human operator, and the red boxes indicate the areas that are not covered.}
        \label{fig:results}
\end{figure*}

\subsection{Evaluation Metric}

First of all, in this paper, we assume that robots and the operator have the same area coverage range $r=6\text{m}$. So, since our goal is strategically positioning robots to maximize the coverage area all over the map, we newly introduce a metric for evaluating this factor. Let $\bar{O}$ be the set of occupancy grid cells of the map area scaled by the map resolution and origin. We define the total map area $A$ and the area covered by all robots and the operator $A_\text{cover}$ as follows:

\begin{align}
    A &= \big| \bar{O} \big| \\
    A_\text{cover} &= \Bigg| \bigcup\limits_{\lambda \in \Lambda\cup\{P\}} B_r[\lambda] \Bigg|\text{,} \label{eq:Acover}
\end{align}
where
\begin{equation}
    B_r[\lambda] = \{ o\in \bar{O} | d(o,\lambda) \leq r\}\text{.} 
\end{equation}

We can see from eq. (\ref{eq:Acover}) that $A_\text{cover}$ is the union of the coverage ranges of all robots and the operator. Therefore, the area coverage percentage ($ACP$) is defined as
\begin{equation}
    ACP = \frac{A_\text{cover}}{A} \cdot 100\text{.}
\end{equation}

\subsection{Benchmark Method}

\begin{figure*}
  \begin{center}
  \includegraphics[width=0.9\linewidth]{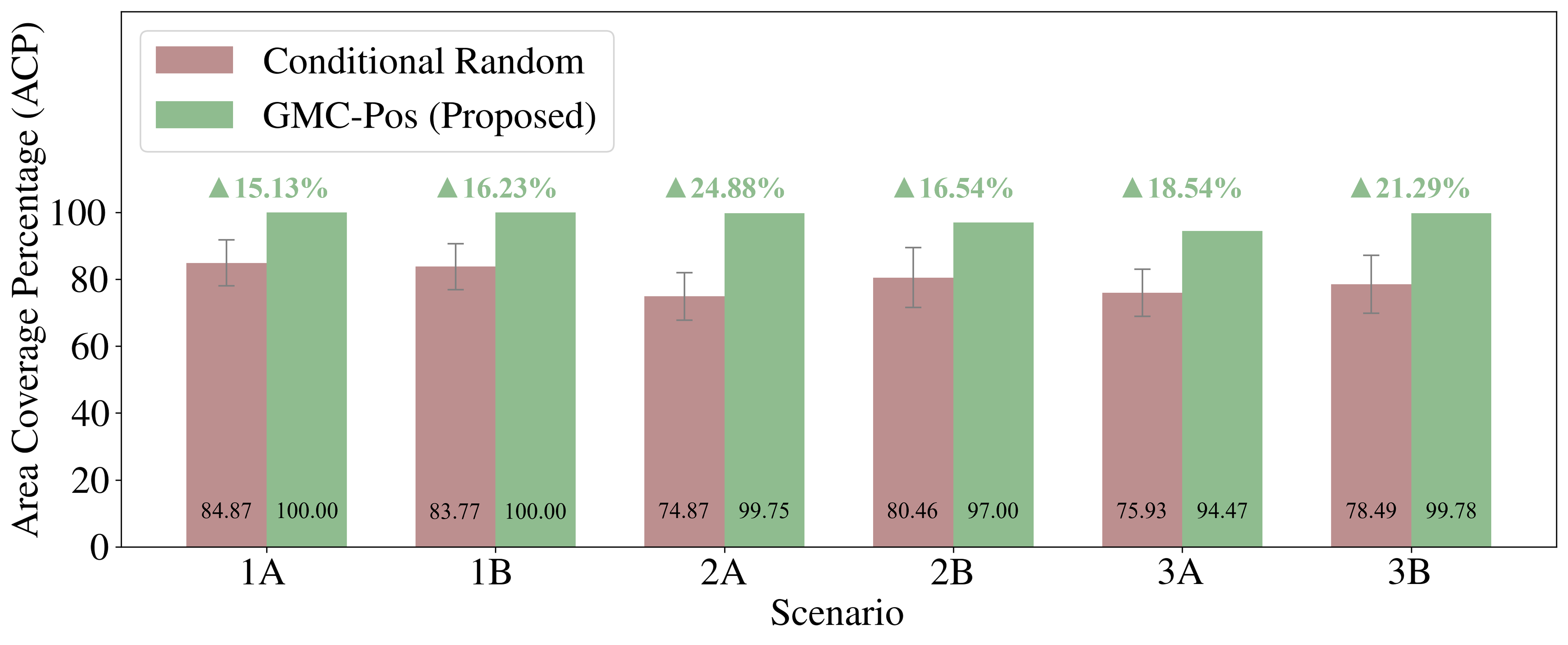}
  \caption{Bar plot for the $ACP$ of the GMC-Pos and the Conditional Random in six scenarios.}
  \label{fig:AC}
  \end{center}
\end{figure*}

Turning now to the benchmark, we implement a method called Conditional Random, as presented in Algorithm \ref{algo:condrand}, to compare with our GMC-Pos, in which by this method, the robot positions will be chosen based on the occupancy grid and the Euclidean distance. Let $O\subseteq \bar{O}$ be the set of unoccupied occupancy grid cells of the map area scaled by the map resolution and origin, $\Psi = \{ \psi_0, \psi_1, ..., \psi_{N-1} \} \subseteq O$ be the set of cells that are selected as the positions for $N$ robots. The selection process is constructed using the recursion as follows:
\begin{align}
    \psi_0 &= \rand(O_p,1) \text{,} \notag \\ \psi_1 &= \rand(O_p,1) \text{,} \notag \\
    \psi_i &= \rand(O_{\psi_{i-2}},1) \quad \notag \\\text{if } i &= 2,3, ..., N-1 \text{,} \label{eq:rand}
\end{align}
where
\begin{equation}
    O_\phi = \{ o\in O | \alpha \leq d(o,\phi) < 2r \} \setminus \Psi\text{.}
\end{equation}
Recall that $d$ is the Euclidean distance, $\alpha$ is the same as in eq. (\ref{eq:alpha}), and $\rand(S,k)$ is a function that chooses random $k$ items from the set $S$. Here the process (\ref{eq:rand}) is in a random manner because of the high number of grid cells needed to be considered and filtered under the conditions. Therefore, for the Conditional Random method, we will use the average of 50 iterations of each scenario for comparison in the following section.

\section{Results and Discussion}
\label{sect:IV}

The simulations for the GMC-Pos and the Conditional Random are conducted in the six scenarios we mentioned previously, as presented in Fig. \ref{fig:results}. Note that for the Conditional Random, the figures shown are an iteration of each scenario setting. They are evaluated using the $ACP$, as shown in Fig. \ref{fig:AC}. We can see that the GMC-Pos performs better than the Conditional Random in all the scenarios. 

Firstly, in Scenarios 1A and 1B, shown in Fig. \ref{fig:1A-GMC-Pos} to \ref{fig:1B-CR}, our purpose is to preliminarily check if the GMC-Pos works correctly on Map 1, which is a simple map. We can see that the robots using the GMC-Pos successfully cover the whole area as the node selection does not fail in choosing the three apparently best robot positions. On the other hand, those using the Conditional Random still have some uncovered areas around one of the corners since the robot positions are not selected efficiently using a graph-based approach. So, the results show that the GMC-Pos has 15.13\% and 16.23\% higher $ACP$ than the Conditional Random in Scenarios 1A and 1B, respectively. Secondly, in Scenario 2A, shown in Fig. \ref{fig:2A-GMC-Pos} and \ref{fig:2A-CR}, the robots using the Conditional Random have an unbalanced positioning between the left and right sides of the map. On the other hand, those using the GMC-Pos can spread equally and cover most of the whole area well, resulting in a 24.88\% improvement of $ACP$. Thirdly, in Scenario 2B, shown in Fig. \ref{fig:2B-GMC-Pos} and \ref{fig:2B-CR}, there is a space and unbalanced distribution among the robots using Conditional Random. The main reason is that the Euclidean distance used in the Conditional Random can sometimes be suboptimal. However, GMC-Pos that uses $d_G$ still performs well in this scenario, resulting in a 16.54\% improvement of $ACP$. Fourthly, in Scenario 3A, shown in Fig. \ref{fig:3A-GMC-Pos} and \ref{fig:3A-CR}, although the robot using the Conditional Random can span through the map, the selected robot positions are not good enough to cover the map. In contrast, the coverage by robots using GMC-Pos is almost the whole map, as we can see 18.54\% higher $ACP$. Finally, in Scenario 3B, shown in Fig. \ref{fig:3B-GMC-Pos} and \ref{fig:3B-CR}, we can see that the distances between robots using the Conditional Random is unbalanced. Consequently, the coverage is not as high as the ones using the GMC-Pos, which almost perfectly covers all parts of the maps, resulting in a 21.29\% higher $ACP$. Moreover, we observe another advantage of the GMC-Pos, which is all the selected positions are located in accessible and practical areas, i.e., they are quite visible to the human operator and not too close to the wall.

\section{Conclusion and Future Work}
\label{sect:V}

This paper proposes the GMC-Pos, a novel positioning method for multiple robots to maximize the environment area coverage. Our approach consists of two modules. Firstly, the graph generation module is for representing the environment map in a practical structure using a connected graph. All the graph nodes are in the accessible area, and the weighted edges show the connectivity and structure of the environment. Secondly, the node selection module is for strategically choosing appropriate positions for robots. We newly introduce the generalized graph-based distance, which combines the Euclidean distance and Dijkstra’s shortest path length. Also, our selection process is based on the recursion with some conditions based on the maximum node degree and the mentioned distance to ensure the chosen positions have the highest area coverage possible. For the simulation, we have six scenarios: multi-robot simulations in a simple map of size 12.20m$\times$12.20m and two challenging maps of size 27.10m$\times$32.20m and 37.37m$\times$23.38m. We compare the positioning performance between our proposed GMC-Pos and the Conditional Random method. The results show that our approach is more well-performed against the Conditional Random regarding the area coverage percentage in all scenarios. There can be further studies to extend the GMC-Pos for making the robots reposition to maintain the area coverage according to the position of the moving human operator. It is also interesting to make physical obstructions such as dense walls and furniture exert influence on the coverage range. Moreover, applying the GMC-Pos for the SAR is a useful and possible topic for future work.

\vspace{0.5cm}

\balance
\bibliographystyle{IEEEtran}
\bibliography{IEEEabrv,Bibliography}

\end{document}